\pdfoutput=1

\documentclass[11pt]{article}

\usepackage{naacl2021}

\usepackage{times}
\usepackage{latexsym}
\usepackage{graphicx}
\usepackage[T1]{fontenc}

\usepackage[utf8]{inputenc}

\usepackage{microtype}

\usepackage{booktabs}
\usepackage[flushleft]{threeparttable}

\definecolor{meixing}{rgb}{0.9, 0.17, 0.31}

\title{Room to Grow: Understanding Personal Characteristics Behind Self Improvement Using Social Media}

\author{MeiXing Dong, Xueming Xu, Yiwei Zhang, Ian Stewart, Rada Mihalcea \\
         University of Michigan, Ann Arbor, MI, USA \\
         \texttt{\{meixingd,xueming,yiweizh,ianbstew,mihalcea\}@umich.edu}}

\begin{document}
\maketitle
\begin{abstract}
    Many people aim for change, but not everyone succeeds. While there are a number of social psychology theories that propose motivation-related characteristics of those who persist with change, few computational studies have explored the motivational stage of personal change. In this paper, we investigate a new dataset consisting of the writings of people who manifest intention to change, some of whom persist while others do not. Using a variety of linguistic analysis techniques, we first examine the writing patterns that distinguish the two groups of people. Persistent people tend to reference more topics related to long-term self-improvement and use a more complicated writing style. Drawing on these consistent differences, we build a classifier that can reliably identify the people more likely to persist, based on their language. Our experiments provide new insights into the motivation-related behavior of people who persist with their intention to change.
    
\end{abstract}

\section{Introduction}
Many people aim for personal change at different points in their lives \cite{baranski2017lay}.
A glance at a list of top-selling books readily yields self-help manuals whose content ranges from implicitly motivating (``Seven Habits of Highly Effective People'' \cite{covey20207}) to explicitly calling for action (``Lean In'' \cite{sandberg2014}). However, simply wanting change is not sufficient to achieve change. Persistence through the process of pursuing personal change is important for actual change to happen, and changes rarely happen overnight. Often, research on behavior change focuses on understanding what makes people committed to regular or increased action, such as exercise \cite{marcus2000physical},
or refraining from certain actions such as not overeating \cite{pappa2017factors}
or not smoking \cite{kanner1999effects}.
An ever-growing number of technological tools, such as food diary apps and wearable activity trackers, have emerged to help monitor and motivate healthy behavior \cite{achananuparp2018does,chung2017personal}.
Regardless of the tools that they use, if someone is not ready for change yet, the intervention is likely to fail \cite{prochaska1997transtheoretical}.

Stage-based models of intentional behavior change posit that people progress through a sequence of two stages \cite{prochaska1997transtheoretical,schwarzer2000social}: \textit{motivation} and \textit{volition}. 
In the initial \textit{motivation} stage, a person develops an intention or goal to act. A person's intention to adopt better behavior depends on factors such as: \textit{risk perceptions}, or the belief that one is at risk of a negative outcome (e.g. ``If I keep procrastinating, I'll fail all my classes."); \textit{outcome expectancies}, or the belief that behavioral change would improve the outcome (e.g. ``If I can have a more consistent daily routine, I will be more successful at work.''); and \textit{perceived self-efficacy}, or the belief that one is capable of doing the desired actions.

In this paper, we seek to understand the characteristics of people who are in the \textit{motivation} stage of behavior change, and how they talk about behavior change. Traditional behavior change tactics focus on convincing people to take action without consideration for what happens during the lead up period \cite{prochaska1997transtheoretical}. Insight into how people act during these earlier stages can help us better understand their needs and inform interventions, such as recommending social media content that exemplifies healthy approaches to self-improvement. They can also help predict later behavior and persistence using early signals. 

\subsection{Research Questions}
We explore how we can computationally model change-seeking behavior and distinguish between those who maintain persistent interest in personal change during the \textit{motivation} phase and those who do not. 
People often turn to social media to express their thoughts and emotions, which provides a rich data source for studying their perceptions and thoughts \cite{dong2019perceptions}. 

We address our research questions using a dataset consisting of the writings of 536 people from an online community focused on self-improvement (the Reddit community r/getdisciplined). In this dataset, we identify those who post frequently and those who post infrequently to identify persistent and non-persistent commitment to change. We analyze the topics, linguistic style, and expressed emotions of the posts authored by the persistent and non-persistent groups of people. Specifically, in this paper, we address three main research questions:

\begin{enumerate}
    \item What are the aspects of life that people want to improve?
    \item What linguistic style do people use to signal their persistent interest in self-improvement?
    \item How does persistent interest in self-improvement reflect in the emotions that authors express?
\end{enumerate}

Using the features tested in the three separate analyses, we are able to classify persistently and non-persistently active authors with over 60\% accuracy, even when using the posts that authors write prior to joining r/getdisciplined.
Considering both the descriptive and predictive analyses, our findings indicate that persistent interest in change can be signalled by early changes in behavior in online discussions.

\section{Related Work}
\textbf{Behavior Change.}
Personal and behavioral change have a long history in the field of psychology \cite{prochaska1994transtheoretical}. 
Improving health behaviors motivated much work in areas like smoking cessation and increasing physical activity.
However, work on understanding how to encourage positive change has expanded to cover countless areas, like decreasing crime \cite{laub2001understanding}, increasing environmentally friendly behavior \cite{semenza2008public}, and enhancing overall well-being \cite{bentley2013health}. This previous work has shown that many factors can influence an intervention's efficacy, a person's willingness to change, and which strategy to choose for a given person \cite{diclemente1998toward}. Further, an intervention's efficacy may change based on where a person is in their process of change. Different stages in the process can be correlated with different levels of attitudes, such as risk perception or self-efficacy \cite{schwarzer2008modeling}. Such attitudes capture a person's estimate of their ability to perform and succeed in challenging situations and are often reflected in the actions that people choose to take or not to take later in later stages \cite{Claro8664,dweck2006mindset,velicer1990relapse}. Several theories of behavioral change delineate stages of change and advocate for interventions tailored to each stage \cite{prochaska1997transtheoretical,schwarzer2008modeling}. 

\textbf{Self-Improvement in Online Communities.}
In recent years, many have turned towards online communities and platforms, such as Reddit and Facebook, to help them make positive personal changes. 
The anonymity available in online discussions helps combat fears of stigma or lack of understanding \cite{ammari2019self}. 
This relative freedom of expression enables researchers to analyze how people seek help through online channels and what they seek \cite{jurgens2015analysis}. People join online communities to obtain support from those with similar experiences \cite{chung2014social}, to ask for guidance and resources \cite{white2001receiving}, and to seek accountability \cite{kummervold2002social}. 
Such support can lead to higher perceived self-efficacy \cite{turner1983social}. 

However, as noted by prior work in behavior change, the type of help needed can be highly dependent on one's personal characteristics and situation. 
In our work, we seek to better understand this using Reddit.
There has been considerable effort spent on learning about people's demographic attributes from social media posts \cite{an2016greysanatomy}.
Work has also targeted internal attributes, such as personality and value, which can be more difficult to extract but can provide richer features for downstream tasks \cite{Yiting}. 
However, few have studied general intentional personal change efforts based on social media posts. We tackle uncovering the underlying linguistic characteristics of those who maintain persistent interest in self-improvement.

\section{Data}
We focus on a Reddit community called r/getdisciplined, where people seek and give advice about how to achieve life goals and build better habits. This community boasts over 768,000 members as of March 2021 and is one of the largest self-improvement subreddits on Reddit. Whereas most self-improvement groups target specific behaviors or goals, such as exercising, losing weight, dieting, or improving mental health, this subreddit targets improving general mental habits. For instance, people ask questions such as ``How do I relearn doing things just for fun?" and ``How do I stop caring about people and craving their attention?" as opposed to questions that are more specific to activities like ``Tips for increasing strength in arms?" or ``How do I eat properly?" 

Each submission, or original post that is not a comment, must designate the intent of the post using a set of specific tags. One can seek advice ([NeedAdvice], [Question]), give advice ([Advice], [Method]), facilitate discussion and accountability ([Discussion], [Plan]), or talk about r/getdisciplined overall ([Meta]). Most submissions seek advice from the community and tend to discuss fundamental issues such as procrastination, lack of motivation, and time management. A sample of submissions are shown in Table \ref{tab:sample_posts}.

From the submissions, we can see clear distinctions between people who seek help. In the first submission, the author expresses that they think a negative trait, procrastination, is probably a set part of their personality and that they do not believe in themselves, resulting in expression of negative emotions (``mildly depressed''). On the other hand, the second submission seeking advice does not make any self-deprecating statements and asks only for productivity tips (``producing quality work''). This implies that they believe in their ability to change their habits with guidance. Across all submissions, it is clear that the writers have made concerted efforts to understand their own behavior.

\begin{table*}[t]
    \centering
    
    \scalebox{0.95}{
    \begin{tabular}{p{15cm}}
         Post \\
        \toprule
        I am a chronic procrastinator without any hope... do you know any drastic measures that might help me turn my life around? I have been procrastinating intensely for pretty much my whole life. It just seems to be a part of my personality at this point. I tried many things but I could never handle it. I have been mildly depressed for a long time now and have no belief in myself whatsoever. \\
        \midrule
        How do you balance Parkinson's Law with producing quality work? I often find myself spending a lot of time on tasks, and I recently read about Parkinson's Law from Tim Ferriss' 4 Hour Workweek. The law states that a project or task will expand to fill the time you have allotted to it. It obviously takes a lot of time and hard work to produce something of quality, whether it be music, writing, etc. How do you stave off Parkinson's Law while still producing something of quality? \\
    \hline
    \end{tabular}
    }
    \caption{Sample [NeedAdvice] posts from the r/getdisciplined subreddit.}
    \label{tab:sample_posts}
\end{table*}

We focus on people who join r/getdisciplined and then become active during a period of five months, from 2017/1 to 2017/5. These are people who had an initial intent to change which turned into continued engagement and persistent intent.\footnote{Data collected using \url{http://pushshift.io}} We categorize people as persistently active in the subreddit, or \textit{persistent}, if they have posted at least four or more times in the given five months.\footnote{This number of posts is the 90th percentile among people who posted during this time.} Only people who have posted in three unique months before and after the target period, respectively, are considered. This pre-processing ensures that there is sufficient data for analysis before, during, and after each person's participation in r/getdisciplined. We then randomly sample an equal number of \textit{non-persistent} people, or people who have posted only once in the 5 months, with the same requirement for posts before and after. Table \ref{tab:data_stats} shows the number of users and posts in our dataset. The total number of users, including both persistent users and a random sample of non-persistent users, is 536.

\begin{table}[]
    \centering
    \scalebox{0.95}{
    \begin{tabular}{l|l}
        Data Summary & \\
        \toprule
        Total Number of Users & 536\\
        Posts from r/getdisciplined & 6010\\
        Posts from other subreddits & 336455 \\
    \end{tabular}
    }
    \caption{Summary statistics about the dataset, such as the number of users and posts.}
    \label{tab:data_stats}
\end{table}

\section{Characteristics of Persistent Interest in Change}
    
    We address the study's questions about persistence in personal change by analyzing the discussed topics, the linguistic style, and the expressed emotions in Reddit posts. We analyze both their general behavior on Reddit \emph{prior} to joining r/getdisciplined as well as their \emph{initial behavior} within r/getdisciplined. Investigating how people act before joining r/getdisciplined helps us learn about the mental or behavioral patterns that indicate a higher likelihood of their intent to change their behavior. As a complement to prior behavior, an individual’s first post indicates how they are approaching behavior change. 
    
    \subsection{What Are the Aspects of Life that People Want to Improve?}\label{sec:topics}
        We uncover the particular areas of life that people seek to improve and their prevalence in discussion. 
        We use topic modeling techniques to uncover the areas of interest that people discuss in their online posts, both within and outside of the context of personal change.
   
        \textbf{Participation in Subreddits.}
            The subreddits, or Reddit communities, in which a person posts shows the general topics with which they engage. We therefore calculate how frequently each user posts in every subreddit, considering only the subreddits that receive 10 posts in aggregate by users that we observe. We consider only the posts made by the users before their first post in r/getdisciplined.
            
            We show the top 50 subreddits for persistent and non-persistent users prior to joining r/getdisciplined in Table \ref{tab:top_subreddits}. We can see that persistent individuals are active in a number of topic-specific self-improvement subreddits, such as \textbf{Fitness}, \textbf{LifeProTips}, and \textbf{personalfinance}. Non-persistent individuals participate in many more gaming subreddits, i.e. related to leisure rather than self-improvement. Both groups post in popular subreddits like \textbf{AskMen}, \textbf{AskReddit}, and \textbf{funny}; the prevalence of ``ask-X'' related subreddits suggests a level of open-mindedness to change that one would expect of people potentially committed to change.
        
            \begin{table*}[]
                \centering
                
                \scalebox{0.9}{
                \begin{tabular}{p{2.5cm}| p{13cm}}
                    User type & Top Subreddits \\
                    \toprule
                    Persistent & Advice, DotA2, EliteDangerous, \textbf{Fitness}, GameStop, GlobalOffensiveTrade, \textbf{LifeProTips}, MakeupRehab, MarvelPuzzleQuest, RWBY, argentina, aww, conspiracy, cowboys, explainlikeimfive, fantasyfootball, hearthstone, me\_irl, \textbf{personalfinance}, photography, relationships, summonerschool, wow \\
                    \midrule
                    Non-persistent & BigBrother, CFB, CringeAnarchy, DeadBedrooms, HelloInternet, IAmA, NoMansSkyTheGame, NoStupidQuestions, OutreachHPG, Roadcam, SquaredCircle, SubredditDrama, WTF, Warframe, baseball, bjj, cars, casualiama, nottheonion, skyrimmods, skyrimrequiem, slatestarcodex, smashbros \\
                    \midrule
                    Both & AdviceAnimals, AskMen, AskReddit, Jokes, MMA, Overwatch, Showerthoughts, The\_Donald, funny, gaming, gifs, leagueoflegends, mildlyinteresting, movies, nba, news, nfl, pcmasterrace, pics, pokemon, pokemongo, politics, soccer, television, todayilearned, videos, worldnews\\
                    \hline
                \end{tabular}
                }
                \caption{Top 50 subreddits prior to joining r/getdisciplined for persistent and non-persistent users respectively, divided into those that correspond to only one group and both groups. Subreddits relevant to self-improvement are bolded.}
                \label{tab:top_subreddits}
            \end{table*}
        
        \textbf{Topics of General Discourse.}
            To gain further insight into the topics that motivated people engage with, we turn to topic modeling. Latent topics can group concepts that overlap between subreddits and ones that differentiate posts in the same subreddit. We use the Latent Dirichlet Allocation (LDA) model \cite{blei2003latent} to discover topics in our dataset. LDA takes a set of documents, $D$, which each contain a sequence of words, and outputs a set of latent topics that make up the documents. We treat each post as a document $d$ and consider all posts made by our target users in the six months prior to joining r/getdisciplined.
            
            To choose the number of topics for the LDA model, we train models on the general posts made prior to r/getdisciplined with $k=5, 10, 15, 20, 30$ and then manually inspect the resulting topics and their constituent words to evaluate intra-topic coherence and inter-topic separation. To do this, for each value of $k$ we look for resulting topics whose words seemed to primarily be related to one topic, as well as having a lower number of overlapping words between topics. 
            We intentionally keep to a smaller number of topics since we qualitatively found that increasing the number of topics past 30 led to much lower coherence. We choose the 30-topic LDA model for our analysis and later classification experiments. Using the resulting model, we examine the content of user posts pre-r/getdisciplined.

            In Table \ref{tab:topic_analysis_table}, we show a subset of topics and label them through a manual inspection of the top words associated with the topic from the LDA model (e.g. ``school", ``college", and ``classes" correspond to the topic labeled ``Education"). We note the topics that differ significantly across posts made by persistent and non-persistent users before joining r/getdisciplined. We see that persistent users talk more about education, indicating pre-existing interest in a common area of self-improvement. On the other hand, non-persistent users discuss music, politics, and Reddit more, which are general or leisure interests that may be less related to one's personal life. 
            
        \begin{table}[]
        \begin{threeparttable}
        \centering
        \scalebox{0.95}{
        \begin{tabular}{l|l|l|l}
        Feature & P & NP & P-NP \\
        \hline
        \textbf{1st post} & & & \\
         Studying & 0.072 & 0.037 & 0.036* \\
         Routines & 0.114 & 0.085 & 0.028 \\ 
         Productivity & 0.062 & 0.073 & -0.011** \\
         Mental Health & 0.102 & 0.105 & -0.002 \\ 
         Time & 0.165 & 0.118 & 0.047* \\
         Goals & 0.086 & 0.071 & 0.015 \\ 
         Encouragement & 0.021 & 0.049 & -0.028* \\
         Habits & 0.129 & 0.083 & 0.046 \\ 
         Conversation & 0.046 & 0.102 & -0.056* \\
         Work & 0.130 & 0.125 & 0.005 \\ 
        \hline
        \textbf{Prior six months} & & & \\
         Music & 0.092 & 0.093 & -0.002** \\
         Relationships & 0.213 & 0.180 & 0.033 \\ 
         News & 0.147 & 0.148 & -0.001* \\
         Finance & 0.172 & 0.186 & -0.014* \\
         Politics & 0.133 & 0.180 & -0.047** \\
         Gaming & 0.164 & 0.188 & -0.024 \\ 
         Education & 0.228 & 0.189 & 0.039** \\
         Reddit & 0.102 & 0.122 & -0.019** \\
         Automobiles & 0.112 & 0.133 & -0.021* \\
         Family & 0.314 & 0.300 & 0.014 \\ 
        \end{tabular}
        }
        \begin{tablenotes}
            \small
            \item $* - p < 0.05, ** - p < 0.01, *** - p < 0.001$ 
        \end{tablenotes}
        \caption{Mean distributions of topics among posts for persistent (P) and non-persistent (NP) users, as well as the differences between them (P-NP). Statistical significance is determined using a two-sided T-test, with the Benjamini-Hochberg Procedure applied to control for multiple hypotheses testing.}
        \label{tab:topic_analysis_table}
        
        \end{threeparttable}
    \end{table}
    
    \begin{table*}[]
        \centering
        \scalebox{0.95}{
        \begin{tabular}{l|l|l|l|l|l|l}
         & \textbf{1st post} & &  & \textbf{Prior 6 mon.} & & \\
        Feature & P & NP & P-NP & P & NP & P-NP \\
        \hline
        \textit{Linguistic Features} & & & &&&\\
         Readability & -9.800 & 43.002 & -52.802*** & 49.163 & 52.971 & -3.807 \\
         Post Length & 96.276 & 47.522 & 48.754*** & 40.572 & 34.548 & 6.024** \\
        \textit{Emotions} &&& &&&\\
         Anticipation & 0.124 & 0.108 & 0.016 & 0.115 & 0.115 & 0.001 \\
         Disgust & 0.031 & 0.024 & 0.007 & 0.044 & 0.044 & -0.001 \\
         Sadness & 0.045 & 0.042 & 0.003 & 0.067 & 0.066 & 0.002 \\
         Trust & 0.109 & 0.090 & 0.019 & 0.135 & 0.133 & 0.003 \\
         Surprise & 0.032 & 0.033 & -0.000 & 0.054 & 0.054 & -0.000 \\
         Anger & 0.038 & 0.029 & 0.009 & 0.059 & 0.066 & -0.007* \\
         Negative & 0.116 & 0.103 & 0.014 & 0.131 & 0.138 & -0.007 \\
         Joy & 0.060 & 0.062 & -0.002 & 0.098 & 0.095 & 0.003 \\
         Fear & 0.059 & 0.047 & 0.013 & 0.070 & 0.073 & -0.003 \\
         Positive & 0.210 & 0.199 & 0.011 & 0.226 & 0.216 & 0.010 \\
        \end{tabular}
        }
        \begin{tablenotes}
            \small
            \item $* - p < 0.05, ** - p < 0.01, *** - p < 0.001$ 
        \end{tablenotes}
        
        \caption{Mean feature values of linguistic and emotion features in posts from persistent (P) and non-persistent (NP) users, as well as the differences between them (P-NP). Note that the differences for different measures are on different scales. Statistical significance is determined using a two-sided T-test, with the Benjamini-Hochberg Procedure applied to control for multiple hypotheses testing.}
        \label{tab:analysis_table}
    \end{table*}
    
        \textbf{Topics of Interest in Self-improvement.} 
            The topics that people discuss in general on Reddit differ greatly from those that are discussed in a focused subreddit. To hone in on the content specific to r/getdisciplined, we train another 30-topic LDA model using all the posts made in r/getdisciplined  between 2016/1 to 2020/2.  
            
            We represent each initial post with the distribution of topics that it contains, according to this LDA model. In Table \ref{tab:topic_analysis_table}, we again show a subset of topics and note those that differ significantly between the two groups of users. Persistent users discuss studying and academics more than non-persistent users, as well as time and time management, showing interest in longer-term shifts in how to go about their life. Non-persistent users engage in more words of encouragement and conversation, perhaps trying to establish connection with the community to increase the likelihood of helpful responses. They also speak about productivity more than persistent users, which is indicative of asking for straightforward productivity tips to solve immediate problems (e.g. ``What apps can I use to help with work?"), rather than tackling longer-term change (e.g. ``I really want to gain some discipline and self control. I would appreciate advice!").

\subsection{What Linguistic Style Do People Use to Signal their Persistent Interest in Self-Improvement?}\label{sec:linguistic_style}
    Patterns in how people express themselves through language can potentially tell us about how they think. Linguistic style has been shown to reflect numerous behavioral characteristics such as personality \cite{ scherer1979personality}, 
    and intent \cite{pennebaker2011using}. We look at the length of each post, taking the number of words contained in the post as a feature. We also consider each post's readability as defined by its Flesch Reading Ease score \cite{kincaid1975derivation}: higher scores indicate longer average word length and sentence length, which implies more difficulty in reading.  We compute these two scores for each post and use these two values as features in our predictive models. As before, we analyze the posts of persistent and non-persistent users both prior to posting in r/getdisciplined and in their first post in the subreddit.

        \textbf{General Linguistic Style.}
        We show the average post lengths and Flesch Reading Ease scores for the prior posts of persistent and non-persistent users in Table \ref{tab:analysis_table}. Persistent users tend to have longer posts than non-persistent users, which could indicate a more committed writing style (e.g., explaining all necessary details of a situation when posting). In contrast, the two groups' posts do not differ much in readability.

        \textbf{Self-Improvement Linguistic Style.}
        Next, we look at the average post lengths and readability scores of initial posts in r/getdisciplined (Tab. \ref{tab:analysis_table}). 
        In contrast to the pre-joining posts, persistent users write significantly longer posts and lower readability, indicating more complex posts. Initial posts that ask for help without self-deprecation, such as the second post in Table \ref{tab:sample_posts} can include many details about the situation at hand so that others can offer pertinent advice. 
            
\subsection{How Does Persistent Interest in Self-improvement Reflect in the Emotions that Authors Express?}\label{sec:emotional_trends}
    The third research question considers trends in emotional expression among people seeking motivation for change. Emotions can signal attitude towards one's intended behavior change. For instance, someone who believes that success is based on innate ability or who expects that they will fail at difficult tasks will probably shy away from goals that require large effort \cite{hutchinson2008effect}. On the other hand, those who believe success results from hard work or believe in their own ability to tackle challenges may be more persistent in their efforts \cite{strecher1986role}.
    
    To analyze such trends, we use the NRC Emotion Lexicon \cite{Mohammad13,mohammad2010emotions}, which contains English words and their associations with positive and negative sentiment as well as eight basic and prototypical emotions \cite{plutchik1980general}: \textit{anger, fear, anticipation, trust, surprise, sadness, joy, and disgust}. Complex emotions, such as \textit{regret} or \textit{gratitude}, can typically be viewed as combinations of these basic emotions. The lexicon contains 14,182 general domain words, each of which can be linked to multiple emotions.

    \textbf{Emotions in General Discourse.}
        Building on our previous observation about the prevalence of emotional words, we now compare the rate of use among persistent and non-persistent people.
        We compute the total proportion of emotions expressed for each person by averaging the counts of emotion words used across the person's posts.
        Comparing the persistent and non-persistent people, we found that most of the emotions are equally found in posts by both groups.
        However, non-persistent users express more anger in general, which may indicate a tendency to be more easily discouraged when faced with difficulty in everyday situations. 
        
    \textbf{Emotions of Self-improvement.}
        We use the same emotion lexicon to extract the expressed emotions in each initial post to r/disciplined. The expressed emotions in first posts that do not differ significantly between persistent and non-persistent users (Table \ref{tab:analysis_table}). However, we see that there is a general trend among everyone of expressing positive sentiment, anticipation, and trust, which signals that they are hopeful with respect to self-improvement and are open to discussing problems and solutions. There is also negative sentiment, which can indicate dissatisfaction towards their current situation and therefore desire to change.
        
\section{Predicting Persistence in Change}
Our analyses have identified that the people who persist in their self-improvement efforts exhibit consistent linguistic differences in topics, writing style, and emotional expression, versus those who do not persist. As a natural next step, we ask whether we can leverage these characteristics to automatically distinguish between these two groups. We set up a prediction task to determine whether a user is likely to become a persistent or non-persistent user on r/getdisciplined by considering: (1) their language use within six months prior to their initial post on r/getdisciplined; (2) their language use in their first post; and (3) their combined language use within the six months prior and their first post on r/getdisciplined. 

To provide more fine-grained semantic representation of the post language, we also construct word embeddings \cite{mikolov2013distributed} from the text of each post, using word2Vec embeddings pre-trained on news text.\footnote{\url{https://code.google.com/archive/p/word2vec/}} Word embeddings are useful in capturing fine differences between words, such as differences in sentiment valence between similar words (e.g. ``good'' vs. ``great''). For each initial post in r/disciplined, we average the word embeddings of each word in the post to generate a per-post embedding. To represent prior posts, we average the per-post embeddings for all posts of each user from the six months prior to joining r/getdisciplined.  For readability, we also include an aggregate readability score based on a number of different readability metrics, in addition to the Flesch score used earlier.\footnote{\url{https://pypi.org/project/textstat/}}

We compare the performance of classifiers that use different combinations of the linguistic features that we have shown to correlate with persistent behavior. Our task is the binary prediction of whether a user will continue to engage (persistent user) or leave after an initial post (non-persistent user). The experiments are performed using SVM classifiers \cite{cortes1995support} and evaluated using 10-fold cross validation.\footnote{We used the SVM classifier, with default parameters, as applied in Scikit-learn: \url{https://scikit-learn.org}} Since our dataset is balanced, both the random and majority class baselines correspond to an accuracy of 50\%.

\begin{table}
    \centering
    \scalebox{0.95}{
    \begin{tabular}{l|l|l|l|l}
    Features & Acc & Prec & Rec & F1 \\
    \hline
    \textbf{1st post} &&&& \\
      Readability & 0.61 & 0.59 & 0.72 & 0.65 \\
      Post Length & 0.60 & 0.57 & 0.83 & 0.67 \\
      Emotionality & 0.54 & 0.54 & 0.57 & 0.55 \\
      W2V & 0.60 & 0.64 & 0.46 & 0.53 \\
      LDA & 0.58 & 0.59 & 0.53 & 0.56 \\
      \textit{Combined} & 0.62 & 0.59 & 0.79 & 0.67 \\
     \hline
     \textbf{Prior six months} &&&& \\
      Readability & 0.53 & 0.52 & 0.63 & 0.57 \\
      Post Length & 0.56 & 0.56 & 0.57 & 0.57 \\
      Emotionality & 0.54 & 0.54 & 0.49 & 0.51 \\
      W2V & 0.56 & 0.55 & 0.58 & 0.57 \\
      Subreddits & 0.55 & 0.54 & 0.62 & 0.58 \\
      LDA & 0.62 & 0.63 & 0.59 & 0.61 \\
      \textit{Combined} & 0.55 & 0.55 & 0.59 & 0.57 \\
     \midrule
      \textit{All} & 0.61 & 0.58 & 0.77 & 0.66 \\
    \end{tabular}
    }
    \caption{Prediction results for binary classification of persistence in r/getdisciplined. Metrics: accuracy, precision, recall, and F1 score.} 
    \label{tab:classification_results}
\end{table}

We present the results in Table \ref{tab:classification_results}, with classification performance shown for each feature set derived from a user’s prior behavior, their first post in r/getdisciplined, and the combination of all features. 
Using all features, our models are able to achieve an average accuracy of over 60\%.
This shows that people who persist with change can be distinguished from those who do not, even before they commit to change by posting in r/getdisciplined.
That said, the models that use only features from each user's initial post in r/getdisciplined yield the highest performance overall. This is in line with previous work showing that the initial posts that someone makes in a conversation can reliably predict future outcomes, such as whether a debate will derail \cite{zhang2018conversations} or a user will remain loyal to a community \cite{ICWSM1715710}. Moreover, someone's first post encapsulates how they approach self-improvement such as whether they think it is possible or is an insurmountable goal, which is reflected in their language use.

\section{Discussion}
The readability of a user's initial post appears highly indicative of their future engagement level. As shown previously in Section \ref{sec:linguistic_style}, persistent users tend to have lower readability in initial posts than non-persistent users. This could be because they come with the intention of engaging with the subreddit, and therefore devote more time to their introductory post hoping for a similar reaction of engagement from the forum. 
Post length is also a strong signal for our models both when we're considering only each user's first post as well as their prior posts on Reddit. Similar to the readability feature, one possible explanation is the higher engagement with the community through longer posts. Users having longer posts prior to joining r/getdisciplined indicates a more consistently personal style of extensive writing and engagement, and therefore more willingness for self-disclosure.

The emotionality features provided some signal for the model, but were not as helpful as our other features. However, emotionality features derived from the 1st post resulted in higher recall than those derived from the prior six months, which could indicate that there is more expressed through emotion in the 1st post than in general text. 

Prediction performance was consistently high when using word embeddings, which shows that the latent semantic information in embeddings is helpful. However, it is not significantly better than the other top features, indicating that there is room for improvement in representing more subtle linguistic information such as intent or attitude.

Topical content features derived through LDA were among the best performing features for activity from the prior 6 months, while a user's subreddit activity history was less predictive. The subreddits in which someone participates might be too coarse-grained for our task, whereas topic models can better capture the fine-grained behavior that relates to self-improvement and mindset. 

Our results demonstrate how people with persistent interest in personal change act differently from those who do not maintain persistent interest. 
Our analyses showed that those with persistent change intent had higher prior engagement with topics that foster personal change, such as education. This kind of behavior represents a form of \textit{gathering information} related to the intended form of change. Information gathering is an important aspect of a person's reflecting and considering their motivation for potential future change \cite{schwarzer2008modeling}.
In addition to topics, we revealed differences in linguistic style between the two groups of people. Persistent users tended to have longer initial posts with lower readability.

\textbf{Implications for Tailored Interventions}
We can use our findings and further work to tailor behavior change interventions towards people with different characteristics. Those characterized with lower persistence may be in an earlier behavior change stage, necessitating a different approach than those in later stages \cite{diclemente1998toward}. For example, a social intervention could consist of a community moderator, or persistent community member, being paired with a likely non-persistent member (based on language use) to encourage them to stay committed to their goal~\cite{vlahovic2014support}. Alternatively, a community-based intervention system could automatically recommend posts from persistent people, for the non-persistent people to read as a way to learn how to approach change in a healthier way~\cite{cosley2007suggestbot}.

\section{Conclusion}

In this paper, we explored the behavior of users from an online community, r/getdisciplined, as a proxy for measuring persistent intent towards personal change. By analyzing user behavior prior to and immediately after joining the community, we showed quantitative differences between users who sustained intent towards general self-initiated change versus those who did not. Those who have persistent intent tended to engage more with change-oriented topics such as education even prior to expressing explicit intent to change.

We then leveraged these linguistic characteristics to build predictive models that were able to automatically distinguish people who continued engagement in r/getdisciplined and sustained their intent for self-improvement from those who did not continue, even before their first post.

Our results provide actionable insight for research areas that investigate behavior change. Understanding the underlying mechanisms associated with persistence in change can support the development of new approaches to help people change for the better.

\section*{Acknowledgments}
This material is based in part upon work supported by the National Science Foundation (grant \#1815291) and by the John Templeton Foundation (grant \#61156). Any opinions, findings, and conclusions or recommendations expressed in this material are those of the authors and do not necessarily reflect the views of the  National Science Foundation or the John Templeton Foundation. 

\bibliographystyle{ACM-Reference-Format}
\bibliography{references}

\end{document}